\def\year{2022}\relax
\documentclass[letterpaper]{article} 
\usepackage{aaai22}  
\usepackage[dvipsnames]{xcolor}
\usepackage{times}  
\usepackage{helvet}  
\usepackage{courier}  
\usepackage[hyphens]{url}  
\usepackage{graphicx} 
\usepackage{natbib}  
\usepackage{caption} 
\DeclareCaptionStyle{ruled}{labelfont=normalfont,labelsep=colon,strut=off} 
\usepackage{algorithm}
\usepackage{algorithmic}
\usepackage{amsmath}
\usepackage{amsfonts}

\usepackage{newfloat}
\usepackage{listings}
\floatstyle{ruled}
\newfloat{listing}{tb}{lst}{}
\floatname{listing}{Listing}
\title{Greenformer: Factorization Toolkit for Efficient Deep Neural Networks}
\author{
    Samuel Cahyawijaya\thanks{The authors contributed equally to this work.}, Genta Indra Winata$^{*}$, Holy Lovenia$^{*}$, Bryan Wilie$^{*}$, \\
    Wenliang Dai, Etsuko Ishii, Elham J. Barezi, Pascale Fung \\
}
\affiliations{
    Center for Artificial Intelligence Research (CAiRE)\\
    The Hong Kong University of Science and Technology\\
    \texttt{scahyawijaya@connect.ust.hk}
}

\usepackage{bibentry}

\begin{document}

\maketitle

\begin{abstract}

While the recent advances in deep neural networks (DNN) bring remarkable success, the computational cost also increases considerably. In this paper, we introduce Greenformer, a toolkit to accelerate the computation of neural networks through matrix factorization while maintaining performance. Greenformer can be easily applied with a single line of code to 
any DNN model. Our experimental results show that Greenformer is effective for a wide range of scenarios. We provide the showcase of Greenformer at \url{https://samuelcahyawijaya.github.io/greenformer-demo/}.

\end{abstract}

\section{Introduction}

With the significant computational growth of DNN models ~\cite{hernandez2020measuring}, AI researchers all around the globe have started to promote and adopt the concept of `Green AI' ~\cite{schwartz2019green}. Many recent works ~\cite{strubell2019energy,lacoste2019quantifying,patterson2021carbon,dai2021mesm,menghani2021efficient} address the environmental challenges such as energy usage and carbon emission level of DNN models and develop more efficient deep learning solutions. In response to this problem, 
we introduce a robust and easy-to-use low-rank matrix factorization toolkit which reduces not only the computational cost but also the model size, with minimal performance loss.

Low-rank matrix factorization is done by decomposing a large matrix into two or more smaller matrices, reducing computation and memory costs. \textit{Post-training factorization} methods with singular-value decomposition (SVD) ~\cite{golub1970svd} and non-negative matrix factorization (NMF) ~\cite{lee2001nmf} have been applied to approximate the weight matrix of a trained model~\cite{winata2019effectiveness,noach2020compressing}. On the other line of work, \textit{factorization-by-design} applies matrix factorization is directly to the model structure prior to the training. This method produces impressive results with the compressed model is not only smaller and faster but also able to outperform the uncompressed model ~\cite{winata2020lrt,cahyawijaya2021greenformers,kuchaiev2017factorization}.


\begin{figure}[t]
\includegraphics[width=\linewidth]{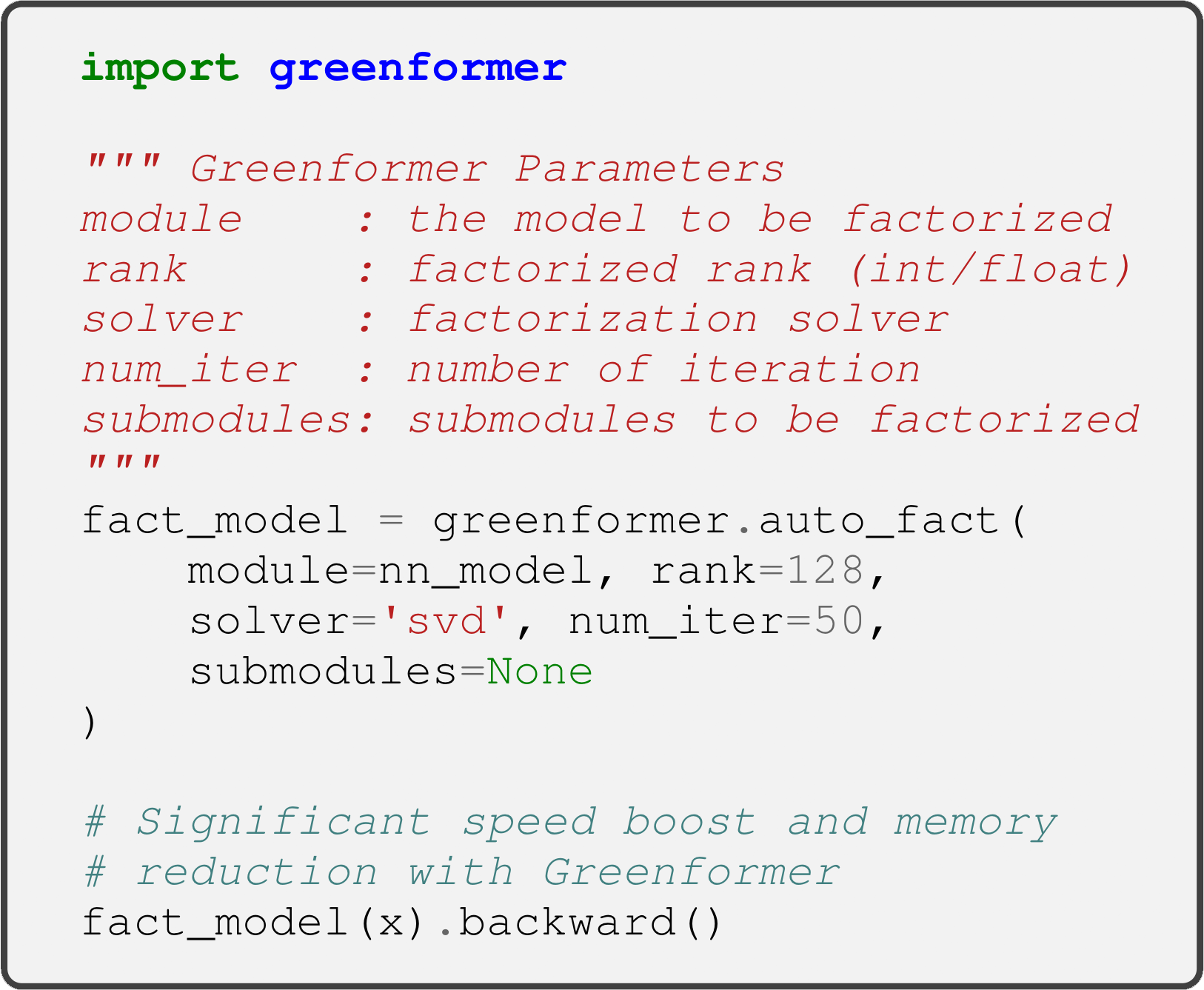}


\caption{Model factorization with Greenformer for an efficient compute time. Greenformer provides efficiency boost with a minimum tweak on the code base.}
\label{fig:pseudo_code}
\end{figure}



Despite the fact that many works have been published on low-rank matrix factorization, all the solutions are model-dependent, making applicability to different model architecture difficult and cumbersome. To improve the generalization and applicability of the low-rank matrix factorization method, we introduce Greenformer, an eloquent low-rank matrix factorization toolkit that supports multiple use cases of matrix factorization and is currently implemented for the PyTorch framework~\cite{paszke2019pytorch}. As shown in Figure ~\ref{fig:pseudo_code}, with Greenformer, we can easily factorize any deep neural networks to perform both factorization-by-design and post-training factorization. We further demonstrate the effectiveness of our Greenformer toolkit for three different use cases: 1) factorization-by-design, 2) post-training factorization, and 3) few-shot via in-context learning factorization.

\begin{figure*}[!t]
    \centering
    \begin{minipage}{.33\linewidth}
        \centering
        \begingroup
        \includegraphics[width=\linewidth]{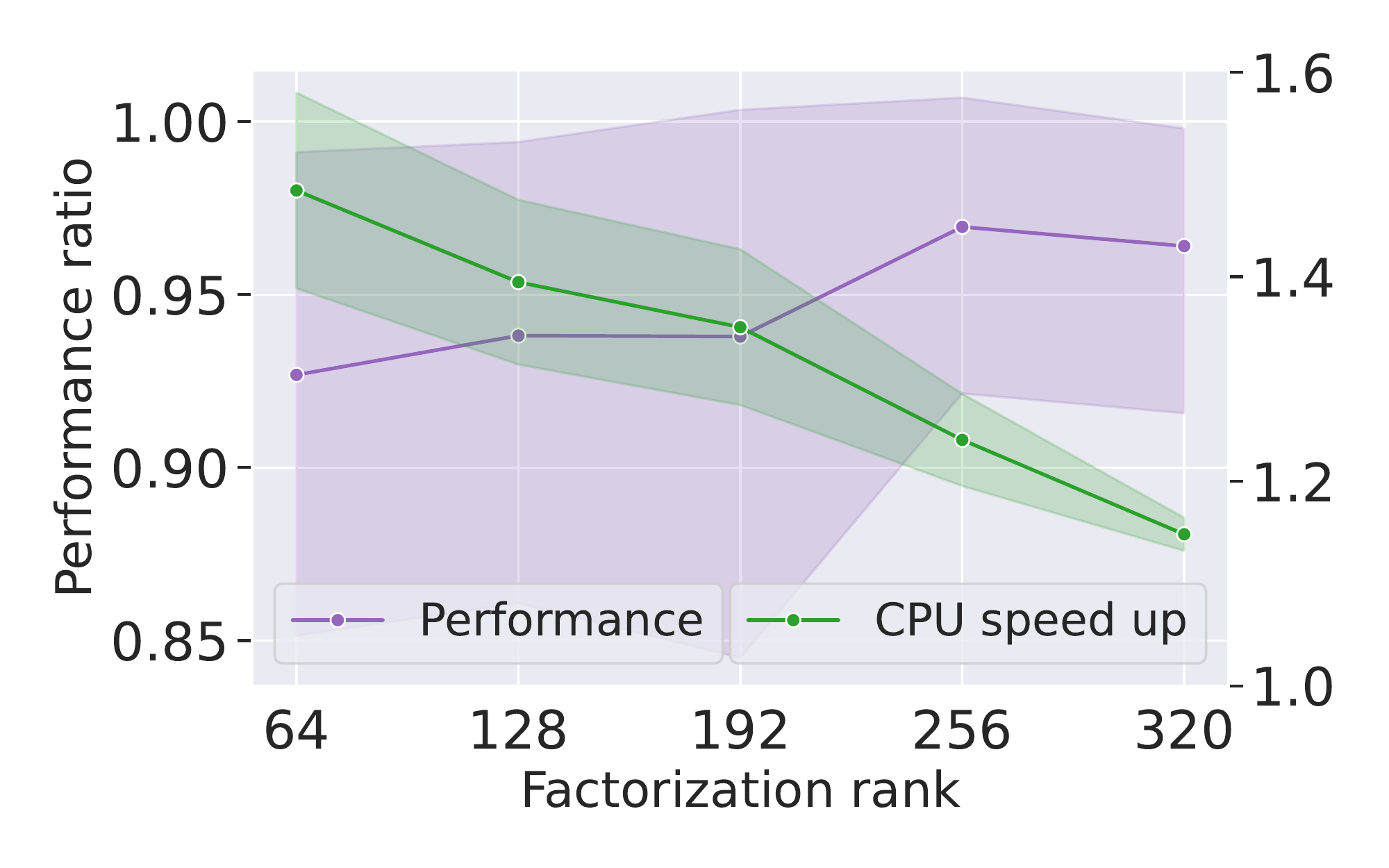}
        \endgroup
    \end{minipage}%
    \begin{minipage}{.33\linewidth}
        \centering
        \begingroup
        \includegraphics[width=\linewidth]{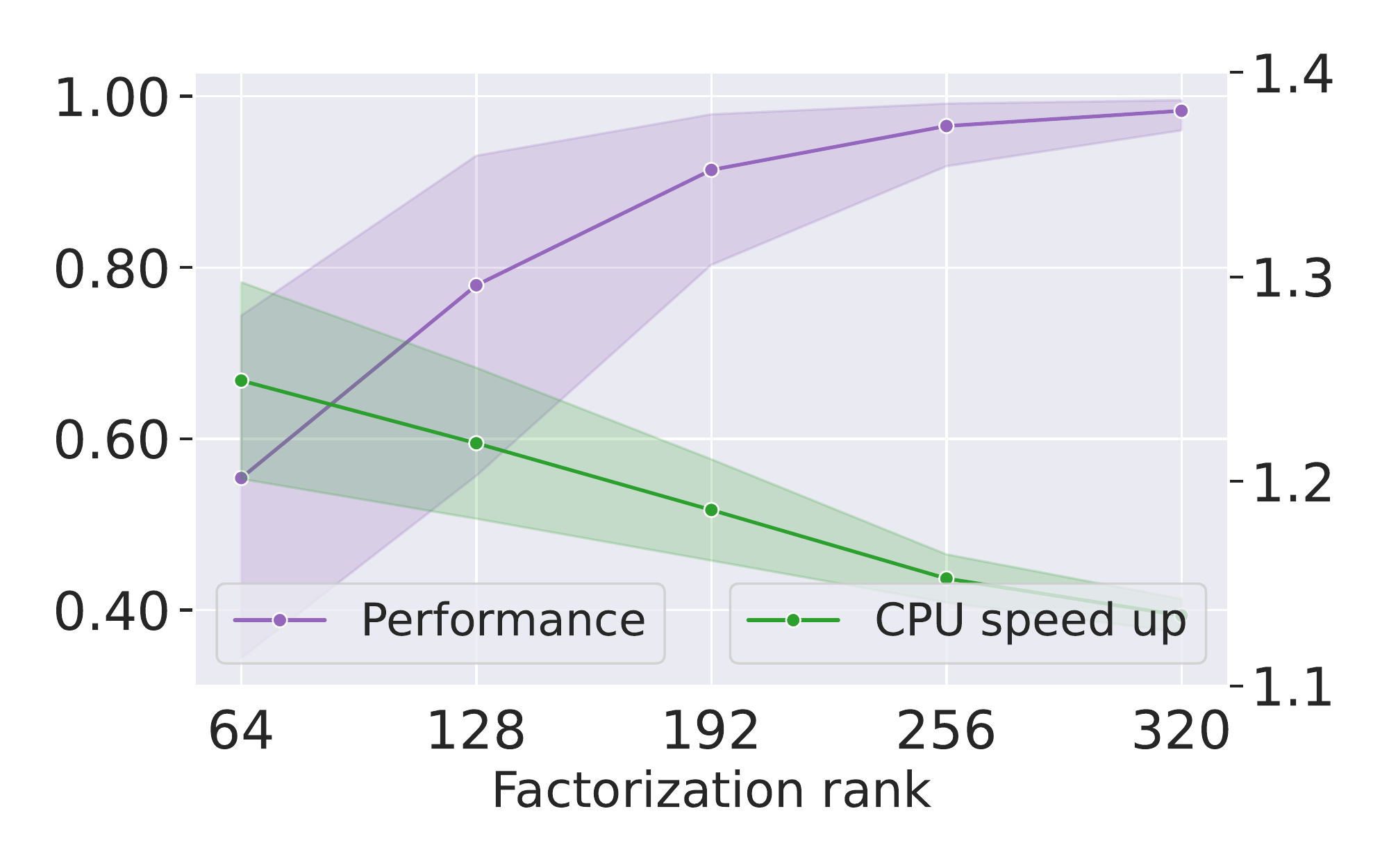}
        \endgroup
    \end{minipage}%
    \begin{minipage}{.33\linewidth}
        \centering
        \begingroup
        \includegraphics[width=\linewidth]{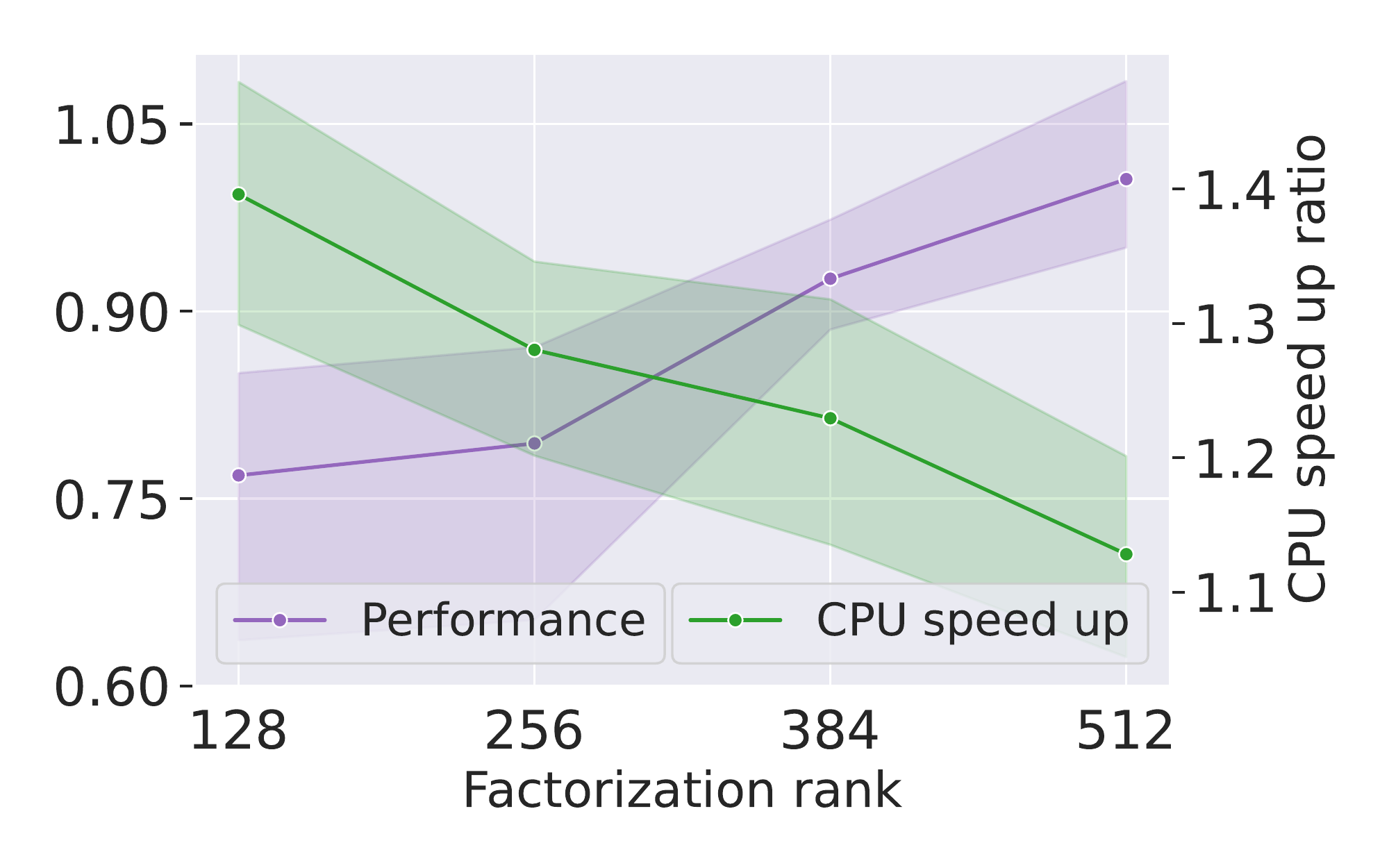}
        \endgroup
    \end{minipage}%
    \caption{Performance and efficiency trade-off of utilizing Greenformer on \textbf{(left)} factorization-by-design, \textbf{(center)} post-training factorization, and \textbf{(right)} in-context learning factorization use cases. \textcolor{Purple}{\textbf{Purple}} and \textcolor{Green}{\textbf{green}} lines denote the relative performance and speed up ratio against the uncompressed model averaged across all tasks.}
    \label{fig:result}
\end{figure*}

\section{Design and Consideration}

\begin{figure}
    \centering
    \includegraphics[width=0.9\linewidth]{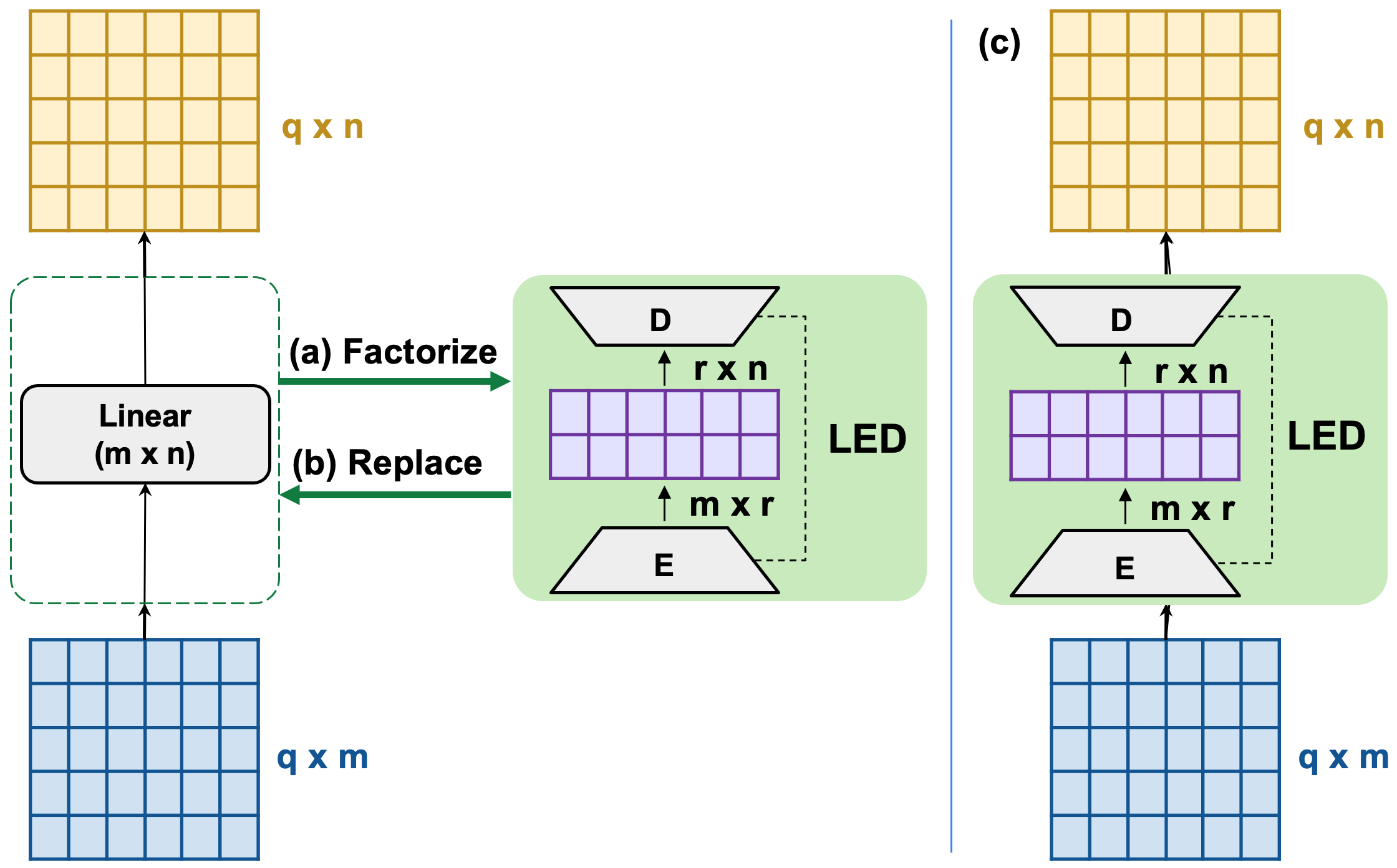}
    \caption{Automatic factorization flow with LED. \textbf{(a)} Linear layer is factorized creating an LED layer.
    \textbf{(b)} The LED layer is used to replace the linear layer in the model producing \textbf{(c)} which requires more efficient than the original linear layer.}
    \label{fig:led_ced}
\end{figure}

Greenformer performs decomposition to the weight matrices of linear and convolution layers. Namely, a weight matrix $W \in \mathbb{R}^{m \times n} $ is decomposed into two low-rank matrices $A \in \mathbb{R}^{m \times r} $ and $B \in \mathbb{R}^{r \times n}$, where $r \ll \min \{m, n \}$.

Greenformer decomposes a matrix by utilizing a factorization solver. There are three different factorization solvers implemented in Greenformer: Random, SVD~\cite{golub1970svd}, and Semi-Nonnegative Matrix Factorization (SNMF)~\cite{lee2001nmf}. Random solver replaces the original matrix with two random matrices by referring the original size and the specified target rank. Note that random solver is not suitable for post-training factorization, since it may break what the model learnt in the main training as it does not approximate the original matrix. SVD solver computes $W = A \Sigma V = AB$ where $\Sigma$ is a diagonal and has singular values. SNMF is an extension of NMF which alleviates the non-negative constraint on $W$. SNMF solver performs decomposition of $W = AB$, where $B$ is strictly non-negative yet $A$ has no restriction on signs. 

As the three solvers mentioned above cannot handle tensors, Greenformer rearranges weight tensors to matrices for decomposition of convolutional layers. For instance, a 1D convolution layer consists of a weight $W \in \mathbb{R}^{C_\mathrm{in} \times C_\mathrm{out} \times S}$, where $C_\mathrm{in}$ and $C_\mathrm{out}$ denote the number of input channel and output channel, and $S$ denotes the size of the convolution kernel. Greenformer rearranges the weight into a 2-dimensional matrix $W' \in \mathbb{R}^{C_\mathrm{in}S \times C_\mathrm{out}}$. The matrix is then decomposed and converted back into the original dimension producing tensors $A \in \mathbb{R}^{C_\mathrm{in} \times r \times S}$ and $B \in \mathbb{R}^{r \times C_\mathrm{out} \times 1}$.
The same trick is also applied for 2D and 3D convolution layers.

The decomposed matrices and/or tensors are then wrapped into a compatible low-rank module which is then used to replace the original linear an/or convolution layers of the model. Specifically, we replace a linear layer into a Linear Encoder-Decoder (LED) layer and replace a convolution layer into a Convolution Encoder-Decoder (CED) layer. The depiction of LED and/or CED layers work is shown in Figure ~\ref{fig:led_ced}. Both LED, and CED have the same input and output with the linear and convolution layers; hence, they can maintain compatibility with the model.

To maximize the outcome of automatic factorization, Greenformer only performs factorization when the low-rank $r$ is less than the maximum low-rank $r_{\max}$ to ensure reduction of the theoretical computational cost. For a given weight matrix $W \in \mathbb{R}^{m \times n}$ the maximum low-rank is defined as:
\begin{align}
r_{\max} &= \frac{(m \cdot n)}{(m + n)}\
\end{align}

To improve its flexibility, Greenformer supports factorization with a dynamic rank across all layers by computing the rank based on a ratio to the maximum rank $r_{\max}$ of the corresponding layer. Additionally, we also observe that applying factorization to all layers of large pretrained models leads to significant performance loss. To address this problem, Greenformer is equipped with a filtering feature that enables factorization only on a specific set of modules. 

We test our toolkit on three use cases: 1) Factorization-by-design, where we train models prior to the training; 2) post-training factorization, where we factorize models prior to evaluation phase; and in-context learning factorization, where we apply factorization to large pretrained language models and perform in-context learning following ~\citet{brown2020language}. We test our toolkit on 3 text classification tasks and 2 image classification tasks. We show the effectiveness of our Greenformer toolkit in all use cases in Figure~\ref{fig:result}.


\section{Conclusion}

We present Greenformer, an automatic factorization toolkit that provides significant efficiency improvement while maintaining the model performance. In addition, Greenformer is flexible, easy-to-use, and applicable for multiple scenarios. For future work, it is interesting to extend Greenformer for more energy-intensive use cases, such as on large models pretraining and network architecture search.



 




\bibliography{aaai22}

\end{document}